\documentclass{article}
\usepackage{ijcai22}

\usepackage{times}

\usepackage{soul}
\usepackage{url}
\usepackage[hidelinks]{hyperref}
\usepackage[utf8]{inputenc}
\usepackage[small]{caption}
\usepackage{graphicx}
\usepackage{amsmath}
\usepackage{booktabs}
\usepackage{amssymb}
\usepackage{makecell}
\usepackage{multirow}
\usepackage{comment}
\usepackage{color}
\title{A Comprehensive Survey on Data-Efficient GANs in Image Generation}

\author{
Ziqiang Li\footnote{Interns at JD Explore Academy}$^1$\and
Beihao Xia$^*$$^2$\and
Jing Zhang$^3$   \and
Chaoyue Wang\footnote{Contact Author}$^4$\and
Bin Li$^\dag$$^1$\\
\affiliations
$^1$University of Science and Technology of China\\
$^2$Huazhong University of Science and Technology\\
$^3$The University of Sydney\quad
$^4$JD Explore Academy\\
\emails
iceli@mail.ustc.edu.cn,
xbh\_hust@hust.edu.cn,
jing.zhang1@sydney.edu.au,
chaoyue.wang@outlook.com,
binli@ustc.edu.cn
}

\begin{document}

\maketitle
\begin{abstract}
	Generative Adversarial Networks (GANs) have achieved remarkable achievements in image synthesis. These success of GANs relies on large scale datasets, requiring too much cost. With limited training data, how to stable the training process of GANs and generate realistic images have attracted more attention. The challenges of Data-Efficient GANs (DE-GANs) mainly arise from three aspects: \emph{(i) Mismatch Between Training and Target Distributions, (ii) Overfitting of the Discriminator, and (iii) Imbalance Between Latent and Data Spaces.} Although many augmentation and pre-training strategies have been proposed to alleviate these issues, there lacks a systematic survey to summarize the properties, challenges, and solutions of DE-GANs. In this paper, we revisit and define DE-GANs from the perspective of distribution optimization. We conclude and analyze the challenges of DE-GANs. Meanwhile, we propose a taxonomy, which classifies the existing methods into three categories: \textbf{Data Selection, GANs Optimization, and Knowledge Sharing}. Last but not the least, we attempt to highlight the current problems and the future directions.
\end{abstract}
\section{Introduction}
Generative Adversarial Networks (GANs)~\cite{goodfellow2014generative} aim to generate images~\cite{karras2020analyzing} indistinguishable from real ones, which have acquired remarkable achievements in computer vision~\cite{wang2018perceptual,tao2019resattr}, image processing~\cite{wang2017tag,liu2019multistage}, and multimodal tasks~\cite{qiao2019mirrorgan,qiao2019learn,chen2020puppeteergan,wu2021f3a}. However, most GANs are trained on large-scale datasets~\cite{brock2018large}, which requires much cost on data collection and model training. For instance, current datasets used to train GANs are FFHQ (70K images), CelebA (200K), and LSUN (1M). In some practical applications, the available  training data is limited. GANs are expected to be trained with limited, few-shot, or even one-shot data (\textit{e.g.,} there are just 10 examples per artist in the Artistic-Faces dataset~\cite{yaniv2019face}). 
\begin{figure}
	\setlength{\abovecaptionskip}{0.1cm}
	\setlength{\belowcaptionskip}{-0.3cm}
	\centering
	\includegraphics[scale=0.45]{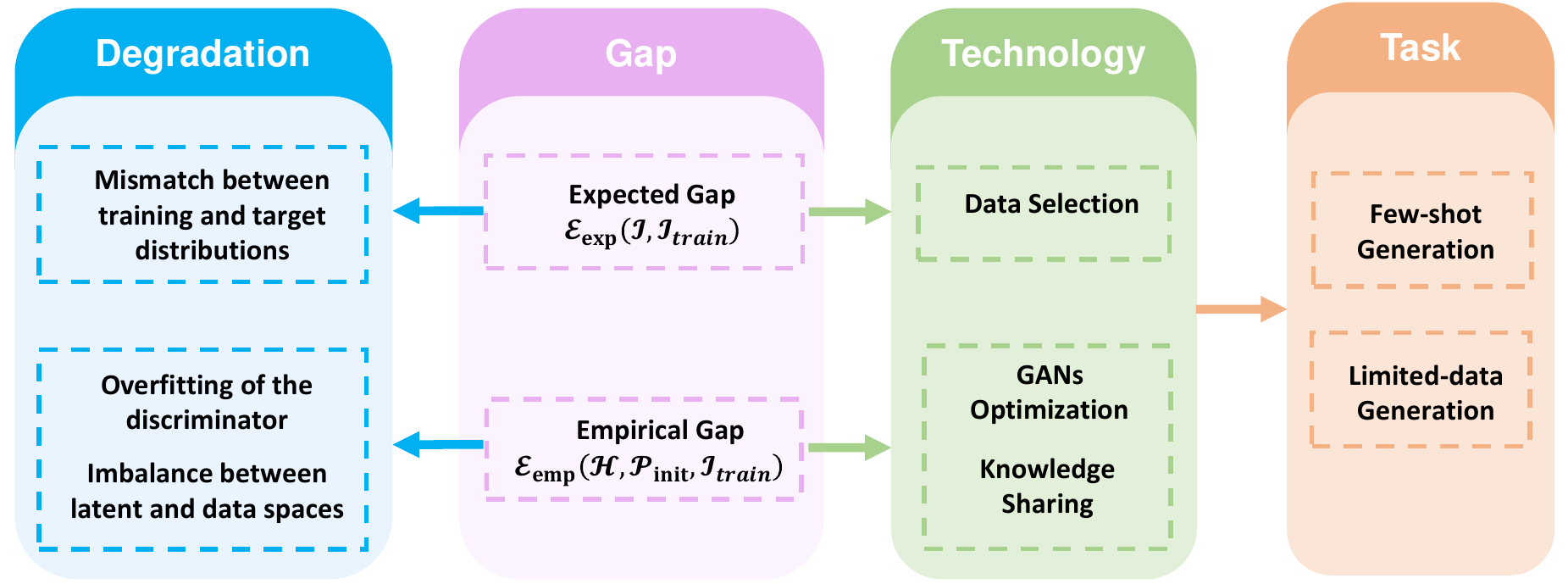}
	\caption{The overview of this survey. This survey first illustrates the target (minimizing gaps) and its upper bound of DE-GANs in Section 2. Then, we demonstrate three challenges (degradation) corresponding to the different gaps of DE-GANs in Section 3. Third, some technologies achieving targets are summarized by a new taxonomy in Section 4. Finally, we divide the task into two parts according to the amount of data in Section 5 and attempt to highlight the current problems and the future directions in Section 6.}
	\label{FIG:flow}
\end{figure}

Although many technologies, such as evolutionary computation~\cite{wang2019evolutionary}, data processing~\cite{li2021high}, and regularization~\cite{gulrajani2017improved,li2020direct} have been proposed for stabilizing GANs training and improving generative performance, they may not perfectly solve the training problems of Data-Efficient GANs (DE-GANs) \cite{karras2020training}. Recently, more and more research works focus on the development of DE-GANs, which aim to fit the entire data distribution using limited training samples~\cite{tseng2021regularizing,yang2021data,karras2020training}.  
These methods attempt to explore the training strategies of DE-GANs from different perspectives, yet a comprehensive survey and related discussions of DE-GANs remain vacant. In this paper, the first taxonomy to summarize technologies in DE-GANs has been presented. The overview of this survey is demonstrated in Figure \ref{FIG:flow}. Specifically, two target (minimizing Data and Empirical gaps) of DE-GANs are defined in Section 2. Then, different challenges (degradation) corresponding to the two targets are described in Section 3. In Section 4, various techniques for minimizing gaps by mitigating degradation of DE-GANs are summarized by a new taxonomy: Data Selection, GANs Optimization, and Knowledge Sharing. Finally, we divide the task into two parts according to the amount of data in Section 5 and conclude the characteristics of datasets, technologies, and evaluation for each task, respectively.

\textbf{The differences between our survey and others.} Many surveys \cite{jabbar2021survey,alqahtani2021applications,gui2021review,wang2021generative,saxena2021generative,li2020systematic} are conducted to summarize applies, architectures, training, loss functions, and regularization of GANs, but they lack a focus on the data efficiency of GANs. Driven by relieving the burden of collecting large-scale high-resolution data in industrial applications and unbearable training time\footnote{StyleGAN-XL \cite{sauer2022stylegan}, setting a new state-of-the-art on large-scale image synthesis, needs 400 V100-days at a resolution of $512^2$ pixels. More seriously, the state of the art diffusion model, ADM \cite{dhariwal2021diffusion}, needs 1914 V100-days at a resolution of $512^2$ pixels. Both of them are intolerable for individuals and academic units.} in academic research, Data-efficient GAN has drawn much attention and is now a hot topic. However, currently, there is no study that provides a comprehensive taxonomy to point out the challenges nor discuss the pros and cons of different approaches in this area. Our survey is the first comprehensive and systematic investigation of the training of Data-Efficient GANs. Furthermore, the survey \cite{wang2020generalizing} also focuses on the Few-Shot Learning problem but seldom mentions the few-shot generation in GANs. To explore the essential issue of data efficiency in GANs, we follow the vanilla GANs setup, the noise-to-image pipeline. Some other settings \cite{gu2021lofgan,clouatre2019figr,hong2020f2gan} of data-efficient generation are not in the scope of this survey.

Our main contributions are summarized as follows:
\begin{itemize}
	\item \textit{Comprehensive analysis on the degradation of DE-GANs.} Three degradation accompanying training of DE-GANs are illustrated in this survey and we provide a comprehensive analysis on these.
	
	\item \textit{New taxonomy.} For the first time, we systematically review and give a formal definition of DE-GANs based on distribution optimization. Accordingly, a novel taxonomy of DE-GANs is proposed:  Data Selection, GANs Optimization, and Knowledge Sharing.
	
	\item \textit{Challenges and opportunities.} Finally, we point out open issues inspiring further research in DE-GANs.
	
\end{itemize}

\section{Problem Definition}

In this section, we provide a formal description of DE-GANs from the perspective of distribution optimization. 
Before defining DE-GANs, we recall the GANs training, which plays a two-player adversarial game between a generator $G$ and a discriminator $D$, where the generator $G(z)$ is a distribution mapping function that transforms low-dimensional latent distribution $p_z$ to target distribution $p_g$. And the discriminator $D(x)$ evaluates the divergence between the generated distribution $p_g$ and real distribution $p_r$. Mathematically, it is formulated as:
\begin{equation}
	\begin{aligned}
		\min _{\phi} \max _{\theta} f(\phi, \theta) =\mathbb{E}_{\mathbf{x} \sim p_{r}}\left[g_{1}\left(D_{\theta}(\mathbf{x})\right)\right] \\
		+\mathbb{E}_{\mathbf{z} \sim p_{z}}\left[g_{2}\left(D_{\theta}\left(G_{\phi}(\mathbf{z})\right)\right)\right],
	\end{aligned}
\end{equation}
where $\phi$ and $\theta$ are parameters of generator $G$ and discriminator $D$, respectively. $g_1$ and $g_2$ are different functions for various GANs. The objective of GANs is to learn a well-trained generator that can fit real distribution well. More perspectives and analyses of GANs have been shown in \cite{li2020systematic}. 

\begin{figure}
	
	\centering
	\includegraphics[scale=0.7]{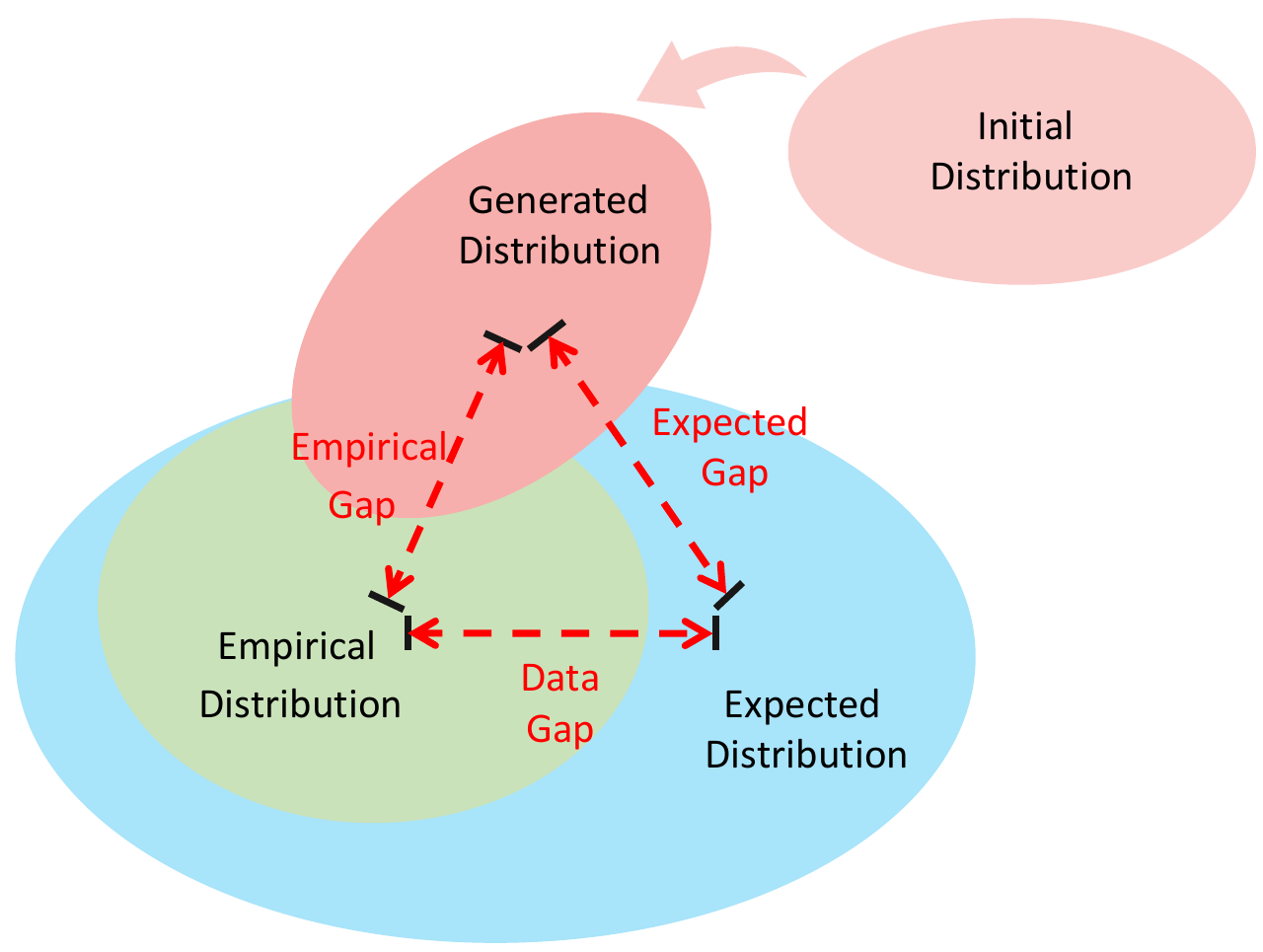}
	\caption{The overview of the DE-GANs. The figure illustrates four distributions and three gaps defined in Section 2.}
	\label{FIG:framework}
\end{figure}

DE-GAN is a special case of GANs, which targets at obtaining a generator that could fit the entire data distribution with limited training samples. 
Here, we provide some definitions of distribution $\mathcal{P}(\cdot)$\footnote{The variables in parentheses of $\mathcal{P}(\cdot)$ and $\mathcal{E}(\cdot)$ represent independent variable. \label{web}}: the whole target distribution is denoted as \textbf{Expected Distribution $\mathcal{P}_{\text{exp}}(\cdot)$}, the limited data distribution used for training is denoted as \textbf{Empirical Distribution $\mathcal{P}_{\text{emp}}(\cdot)$}, and the \textbf{Initial Distribution} and the \textbf{Generated Distribution} are denoted as \textbf{$\mathcal{P}_{\text{init}}(\cdot)$} and \textbf{$\mathcal{P}_{\text{gen}}(\cdot)$}, respectively. Thus, the target of DE-GANs can be represented as minimizing Expected Gap $\mathcal{E}_{\text {exp}}(\mathcal{H},\mathcal{P}_{\text{init}},\mathcal{I},\mathcal{I}_{\text{train}})$, formulating with:

\begin{equation}
	\begin{aligned}
		\mathcal{E}_{\text {exp}}(\mathcal{H},\mathcal{P}_{\text{init}},\mathcal{I},\mathcal{I}_{\text{train}})=M(\mathcal{P}_{\text{exp}}(\mathcal{I}),\mathcal{P}_{\text{gen}}(\mathcal{H},\mathcal{P}_{\text{init}},\mathcal{I}_{\text{train}})),
	\end{aligned}
\end{equation}
where $\mathcal{I}$ is the entire dataset and $\mathcal{I_{\text{train}}}$ represents the limited samples for training DE-GANs ($\mathcal{I_{\text{train}}}\in \mathcal{I}$), $\mathcal{H}$ represents model frameworks and training algorithms, and $M$ measures discrepancy between two distributions. However, it is nearly impossible to directly minimize the $ \mathcal{E}_{\text {exp}}(\mathcal{H},\mathcal{P}_{\text{init}},\mathcal{I},\mathcal{I}_{\text{train}})$ using conventional objectives of GANs due to the unknown of $\mathcal{I}$ during DE-GANs training. Therefore, An upper bound of ${\mathcal{E}_{\text {exp}}(\mathcal{H},\mathcal{P}_{\text{init}},\mathcal{I},\mathcal{I}_{\text{train}})}$ can be defined as:
\begin{equation}
	\begin{aligned}
		\mathcal{E}_{\text {exp}}(\mathcal{H},\mathcal{P}_{\text{init}},\mathcal{I},\mathcal{I}_{\text{train}})\leq \mathcal{E}_{\text {data}}(\mathcal{I},\mathcal{I}_{\text{train}})+\mathcal{E}_{\text {emp}}(\mathcal{H},\mathcal{P}_{\text{init}},\mathcal{I}_{\text{train}}),
	\end{aligned}
\end{equation}
where Data Gap $\mathcal{E}_{\text {data}}(\mathcal{I},\mathcal{I}_{\text{train}})$ is the distance between two data distributions: $\mathcal{P}_{\text{exp}}(\mathcal{I})$ and $\mathcal{P}_{\text{emp}}(\mathcal{I}_{\text{train}})$, which is determined before training and cannot be minimized by optimization. Meanwhile, Empirical Gap $\mathcal{E}_{\text {emp}}(\mathcal{H},\mathcal{P}_{\text{init}},\mathcal{I}_{\text{train}})$ is determined by three factors: model framework and training algorithms $\mathcal{H}$, init generated distribution $\mathcal{P}_{\text{init}}$ (initialization), and limited training dataset $\mathcal{I}_{\text{train}}$. Formally, $\mathcal{E}_{\text {data}}(\mathcal{I},\mathcal{I}_{\text{train}})$ and $\mathcal{E}_{\text {emp}}(\mathcal{H},\mathcal{P}_{\text{init}},\mathcal{I}_{\text{train}})$ are defined as:
\begin{equation}
	\begin{aligned}
		&\mathcal{E}_{\text {data}}(\mathcal{I},\mathcal{I}_{\text{train}})=M(\mathcal{P}_{\text{exp}}(\mathcal{I}),\mathcal{P}_{\text{emp}}(\mathcal{I}_{\text{train}})),\\
		&\mathcal{E}_{\text {emp}}(\mathcal{H},\mathcal{P}_{\text{init}},\mathcal{I}_{\text{train}})=M(\mathcal{P}_{\text{emp}}(\mathcal{I}_{\text{train}}),\mathcal{P}_{\text{gen}}(\mathcal{H},\mathcal{P}_{\text{init}},\mathcal{I}_{\text{train}})).
	\end{aligned}
\end{equation}
Therefore, target of DE-GANs can be realized by minimizing $\mathcal{E}_{\text {data}}(\mathcal{I},\mathcal{I}_{\text{train}})+\mathcal{E}_{\text {emp}}(\mathcal{H},\mathcal{P}_{\text{init}},\mathcal{I}_{\text{train}})$. Figure \ref{FIG:framework} illustrates relationships of four distributions and three gaps.

\section{Degradation of DE-GANs}\label{Sec:challenges}
Minimizing $\mathcal{E}_{\text {data}}(\mathcal{I},\mathcal{I}_{\text{train}})+\mathcal{E}_{\text {emp}}(\mathcal{H},\mathcal{P}_{\text{init}},\mathcal{I}_{\text{train}})$ is still a daunting task due to the degradation of DE-GANs. Specifically, mismatch between training and target distributions may lead to a large expected gap $\mathcal{E}_{\text {data}}(\mathcal{I},\mathcal{I}_{\text{train}})$. When minimizing $\mathcal{E}_{\text {emp}}(\mathcal{H},\mathcal{P}_{\text{init}},\mathcal{I}_{\text{train}})$, overfitting of the discriminator and imbalance between latent and data spaces would be more serious with limited training data. 


\subsection{Mismatch Between Training and Target Distributions}
DE-GANs target at fitting entire data distribution $\mathcal{P}_\text{exp}(\mathcal{I})$ with limited training samples $\mathcal{I}_\text{train}$. There could have a huge gap between training and target distributions posing a great challenge to DE-GANs, inevitably. For selected training samples $\mathcal{I}_\text{train}$ and its represented distribution $\mathcal{P}_\text{emp}(\mathcal{I}_\text{train}$), sparse random sampling does not converge on the $\mathcal{P}_\text{exp}(\mathcal{I})$~\cite{tucker1959generalization} and intensive partial sampling introduces huge domain gap between $\mathcal{P}_\text{exp}(\mathcal{I})$and $\mathcal{P}_\text{emp}(\mathcal{I_\text{train}})$~\cite{gal2021stylegan}. 

\subsection{Overfitting of the Discriminator}
Neural networks (NNs) are easy to overfit when training data is limited, with no exception for the discriminator. ~\cite{zhao2020differentiable,karras2020training} demonstrated that even for large datasets and advanced network architectures, the training of GANs still suffers from severe overfitting. The less training data there is, the earlier the overfitting happens. With only a few training samples, the discriminator tends to remember each of them, showing overly confident and unreasonable decision boundaries. \cite{yang2021data,tseng2021regularizing} also illustrated that the predictions of real and generated images diverge rapidly accompanying the DE-GANs training. \cite{karras2020training} demonstrated that Fréchet Inception Distance (FID) starts to rise at some point during training and outputs of discriminator keep drifting apart during training, especially for DE-GANs, which are strong evidence of the discriminator overfitting in DE-GANs.

\subsection{Imbalance Between Latent Space and Data Space}
Another important factor leading to poor generated results of DE-GANs is the imbalance between latent and data spaces. Latent space obeys Mixed Gaussian Distribution that is \textbf{Continuous} and \textbf{Dense} while data space is \textbf{Discrete} and \textbf{Sparse}. It leads to a dilemma: approximating the discontinuous distribution transform map with continuous NNs~\cite{an2020ae}. 

Ablation studies of \cite{shahbazi2022collapse} illustrated that generator may be the key point of degradation for DE-cGANs (data-efficient conditional GANs) compared to DE-GANs. However, there is no further discussion of latent space in DE-cGANs or DE-GANs. We point out that class-conditioning makes the latent space irrelevant to the category, which further makes the imbalance between latent and image spaces. Furthermore, \cite{kong2021smoothing} showed that DE-GANs display undesirable properties like ``stairlike" latent space where transitions suffer from discontinuity, occasionally yielding abrupt changes in output. It pointed out that this discontinuity of latent space may be the key point in the degradation of DE-GANs. Furthermore, inadvertently used latent space smoothing under discriminator feature metric acquired significant improvement in DE-GANs~\cite{yang2021data}. All of them are evidence of the hazards of imbalance between latent and data spaces for DE-GANs. Although more and more work is beginning to focus on the latent space of DE-GANs, there does not seem to be a consensus in the community on the assumption of "Imbalance Between Latent Space and Data Space".

\section{Taxonomy}
According to different forms of methods in DE-GANs, a new taxonomy denoting as \textbf{``Data Selection"}, \textbf{``GANs Optimization"}, and \textbf{``Knowledge Sharing"} are concluded and discussed in this section. Among them, Data Selection aims to minimize $\mathcal{E}_{\text {data}}(\mathcal{I},\mathcal{I}_{\text{train}})$. Furthermore, GANs Optimization and Knowledge Sharing minimize $\mathcal{E}_{\text {emp}}(\mathcal{H},\mathcal{P}_{\text{init}},\mathcal{I}_{\text{train}})$ by improving optimization strategy on the current limited dataset and by introducing additional pre-training with another large-scale dataset, respectively.
\subsection{Data Selection}

Data Selection is a straightforward strategy to minimize $\mathcal{E}_{\text {data}}(\mathcal{I},\mathcal{I}_{\text{train}})$.  
To learn a better generative distribution $\mathcal{P}_{\text{gen}}$ (\textit{i.e.} be closer to $\mathcal{P}_{\text{exp}}$), the selection of limited samples (\textit{i.e.} $\mathcal{P}_{\text{emp}}$) is critical for DE-GANs~\cite{devries2020instance}. Specifically, a dense sampling reduces the difficulties of DE-GANs training, yet accompanied by a reduction in diversity. Conversely, sparse sampling increases the diversity yet also increases the training instability. For instance, \cite{devries2020instance} removed low-density regions from the data manifold as prior to DE-GANs. It ultimately improved sample fidelity, and significantly reduced training time and model capacity requirements, while also reduced the overall diversity. Although this dense sampling strategy improves the performance of DE-GANs by reducing the training difficulty, it pulls down the top line of DE-GANs, making it impossible to fit the $\mathcal{P}_{\text{exp}}(\mathcal{I})$ through training. Ideally, we should choose the $\mathcal{P}_{\text{emp}}(\mathcal{I}_{\text{train}})$ that is the most available to cover the $\mathcal{P}_{\text{exp}}(\mathcal{I})$. Some studies~\cite{toneva2018empirical,katharopoulos2018not,zhang2021efficient} demonstrated that there exists a small but critical core set from a large dataset that could be hard to learn or be easy to forget by NNs.  These samples are more important for forming the decision boundary of the classifier, by which way a significant fraction of examples can be omitted from training while still maintaining the trained models’ generalization. Accordingly, a similar core set may also be found in GANs training. Apart from choosing $\mathcal{P}_{\text{emp}}(\mathcal{I}_{\text{train}})$ that can represents $\mathcal{P}_{\text{exp}}(\mathcal{I})$ well, some data pre-processing or data augmentation methods can also be applied to make  $\mathcal{P}_{\text{emp}}(\mathcal{I}_{\text{train}})\rightarrow \mathcal{P}_{\text{exp}}(\mathcal{I})$.

\subsection{GANs Optimization}

GANs Optimization adopts some novelty algorithms or frameworks ($\mathcal{H}$) to minimize $\mathcal{E}_{\text {emp}}(\mathcal{H},\mathcal{P}_{\text{init}},\mathcal{I}_{\text{train}})$, which is similar to conventional GANs training. Mathematically, the vanilla form for minimizing $\mathcal{E}_{\text {emp}}(\mathcal{H},\mathcal{P}_{\text{init}},\mathcal{I}_{\text{train}})$ is formulated as:

\begin{equation}
	\begin{aligned}
		\min _{\phi} \max _{\theta} f(\phi, \theta) =\mathbb{E}_{\mathbf{x} \sim \mathcal{P}_{\text{emp}}(\cdot)}\left[g_{1}\left(D_{\theta}(\mathbf{x})\right)\right] \\
		+\mathbb{E}_{\mathbf{z} \sim p_{z}}\left[g_{2}\left(D_{\theta}\left(G_{\phi}(\mathbf{z})\right)\right)\right], 
	\end{aligned}
\end{equation}
which is similar to the formula of conventional GANs in Eq (1). In conventional GANs, we usually assume that the generator trained using large-scale data can fit the given empirical distribution~\cite{karras2020analyzing}. However, minimizing $\mathcal{E}_{\text {emp}}(\mathcal{H},\mathcal{P}_{\text{init}},\mathcal{I}_{\text{train}})$ in DE-GANs is a challenge task~\cite{tseng2021regularizing} due to overfitting of the discriminator and imbalance between latent and data spaces with limited training data. Therefore, some additional data augmentation and regularization technologies are proposed to improve the stability of optimization. Moreover, some simple and efficient architectures have been proposed since complex architectures lead to overfitting with limited training data~\cite{chen2021data}.

\subsubsection{Data Augmentation}
Data augmentation is a popular technology to mitigate the overfitting of NNs \cite{ying2019overview,shorten2019survey} and adversarial robust \cite{rice2020overfitting,rebuffi2021data}. A variety of data augmentation strategies have been applied to the training of GANs, especially DE-GANs. Different from classical data augmentation, augmentation of GANs~\cite{zhao2020differentiable,karras2020training,zhao2020image,tran2021data,jiang2021deceive} augment both real and fake samples and let gradients propagate through the augmented samples to $G$, formulating as:
\begin{equation}
	\begin{aligned}
		\min _{\phi} \max _{\theta} f(\phi, \theta) =\mathbb{E}_{\mathbf{x} \sim \mathcal{P}_{\text{emp}}}\left[g_{1}\left(D_{\theta}\left(T(\mathbf{x})\right)\right)\right] \\
		+\mathbb{E}_{\mathbf{z} \sim p_{z}}\left[g_{2}\left(D_{\theta}\left(T(G_{\phi}(\mathbf{z}))\right)\right)\right],
	\end{aligned}
\end{equation}
where $T(\cdot)$ represents different data transformation technologies of augmentation. According to different forms, transformation can be divided into two types. One type is spatial transformation , such as zoomin, cutout, and translation. The other is visual transformation, such as brightness and coloring. \cite{zhao2020image} demonstrates that augmentations resulting in spatial changes improve the GANs performance more than those inducing mostly visual changes. We conjecture that the above conclusions are also valid for DE-GANs, but there is no evidence. As representative of GANs augmentation, Adaptive Discriminator Augmentation (ADA)~\cite{karras2020training} devised an adaptive approach that controls the strength of data augmentations and Adaptive Pseudo Augmentation (APA)~\cite{jiang2021deceive} adaptively selected generated images to augment real data when the training of discriminator suffers from overfitting. Data augmentation is a striking method for mitigating overfitting of the discriminator and orthogonal to other ongoing researches on training, architecture, and regularization. Therefore, popular augmentation strategies, such as ADA, have been employed as default operations in most DE-GANs. Although ADA~\cite{karras2020training} acquires impressive improvement in DE-GANs, it is not conducive to GANs training in datasets with large amounts of data and little diversity \cite{karras2020training}, such as FFHQ.

\subsubsection{Regularization}

Regularization is a popular technology for enhancing generalization in NNs by introducing priors or extra supervision tasks. A variety of regularization technologies have been summarized in GANs training~\cite{li2020systematic}. In particle, most of the regularization proposed in GANs can also be used in DE-GANs. Nevertheless, since DE-GANs face bigger challenges than traditional GANs, some recent studies are specific to limited data settings, which introduce stronger prior and more demanding tasks. In this survey, we divided them into two parts. 

\textit{1) Overfitting of the discriminator:} Discriminator is easy to suffer from overfitting when training data is limited. To mitigate this problem, Tseng \textit{et al.} \cite{tseng2021regularizing} proposed an anchors-based regularization term to push the discriminator to mix the predictions of real and generated images, as opposed to differentiating them, which offers meaningful constraints for optimizing a more robust objective. The term of regularization is:
\begin{equation}
	R_{\mathrm{LC}}=\underset{\boldsymbol{x} \sim \mathcal{P}_{\text{emp}}}{\mathbb{E}}\left[\left\|D(\boldsymbol{x})-\alpha_{F}\right\|^{2}\right]+\underset{\boldsymbol{z} \sim p_{\boldsymbol{z}}}{\mathbb{E}}\left[\left\|D(G(\boldsymbol{z}))-\alpha_{R}\right\|^{2}\right],
\end{equation}
where $\alpha_{F}$ and $\alpha_{R}$ are two anchors to track the discriminator’s predictions of the generated and real images. Yang \textit{et al.} \cite{yang2021data} introduced instance discrimination, a more difficult task than $\text{Real \& Fake}$ classification, for DE-GANs training through contrastive learning. Among them, InsGen \cite{yang2021data} significantly increases the efficiency of the data, which improves the performance of both GANs and DE-GANs.

\textit{2) Imbalance between latent space and data space:} Furthermore, regularization is also used to alleviate the imbalance between latent and data spaces by introducing latent priors. Latent space in GANs is smooth and interpolable ~\cite{shen2020interpreting,li2021interpreting}. However, Kong \textit{et al.} \cite{kong2021smoothing} found unsmooth and discontinue latent space after DE-GANs training. Based on this, they sample interpolation coefficients $c$ from a Dirichlet distribution and generated an anchor point $z_0$ through interpolation. Then they enforce pairwise similarities between intermediate generator activation to follow the interpolation coefficients. The proposed regularization loss can be formulated as:
\begin{equation}
	\begin{aligned}
		\mathcal{L}_{\text {dist }}^{G} &=\mathbb{E}_{z \sim p_{z}(z), \mathbf{c} \sim D i r(\mathbf{1})}\left[D_{K L}\left(q^{l} \| p\right)\right], \\
		q^{l} &=\operatorname{softmax}\left(\left\{\operatorname{sim}\left(G^{l}\left(z_{0}\right), G^{l}\left(z_{i}\right)\right)\right\}_{i=1}^{N}\right), \\
		p &=\operatorname{softmax}\left(\left\{c_{i}\right\}_{i=1}^{N}\right),\\
		z_{0}&=\sum_{i=1}^{N} c_{i} z_{i}, \quad \mathbf{c} \sim \operatorname{Dir}\left(\alpha_{1}, \cdots, \alpha_{N}\right),
	\end{aligned}
\end{equation}
where $\operatorname{sim}(\cdot)$ denotes the cosine similarity between two features. \cite{yang2021data} adopted contrastive learning to limit the diversity of latent space. The loss can be roughly defined as: $\operatorname{sim}\left(G\left(z\right), G\left(z+\epsilon\right)\right)$, where $\epsilon$ stands for the perturbation term, which is sampled from a Gaussian distribution
whose variance is sufficiently smaller than that of $z$. 

\subsubsection{Model Architecture}
Architecture is also a key point to improve the performance and stability of GANs. Recently, style-based~\cite{karras2020analyzing,karras2019style} and progressive growing based~\cite{karras2017progressive} architectures have attracted much attention from the community. Nevertheless, these structures with numerous parameters easily cause overfitting of the discriminator in limited data settings. Therefore, some lightweight networks are employed in DE-GANs. For instance, \cite{liu2020towards} proposed two technologies, a skip-layer channel-wise excitation module and a self-supervised discriminator trained as a feature-encoder, to gain superior quality
on 1024 $\times$ 1024 resolution with a light-weight structure (FastGAN). The proposed FastGAN achieves better performance and faster training than StyleGAN2~\cite{karras2020analyzing} when data and computing budget are limited. In addition, ~\cite{shaham2019singan,sushko2021one,hinz2021improved} proposed multi-stage and multi-resolution training for one-shot generation. The training progresses through several “stages”, at each of which more layers are added to the generator and the image resolution is increased. At each stage, all previously trained stages are frozen and only the newly added layers are trained. This multi-stage strategy significantly stabilizes training and makes it possible to leaning generative model with only one image. 

Furthermore, \cite{chen2021data} searched sparse subnetworks, which are light-weight and data-efficient compared to original discriminator and generator. Training the sparse subnetworks achieved better performance in limited data settings. Jiang \textit{et al.} \cite{cui2021genco} devised a Generative Co-training network (GenCo) that mitigated the discriminator overfitting by introducing multiple complementary discriminators in DE-GANs. The multiple discriminators provided diverse supervision from multiple distinctive views in training. 

\subsection{Knowledge Sharing}

Knowledge Sharing is another method to minimize $\mathcal{E}_{\text {emp}}(\mathcal{H},\mathcal{P}_{\text{init}},\mathcal{I}_{\text{train}})$ through pre-training on a source dataset different from the expected data $\mathcal{I}$, and fine-tuning on target dataset $\mathcal{I}_{\text{train}}$, which reduces data demand and speeds up training effectively. 
Usually, the choice of source datasets is critical. Recently, {\cite{grigoryev2022when}} demonstrated that ImageNet pre-trained GANs appear to be an excellent starting point for fine-tuning in most of target tasks. Furthermore, the choice of pre-trained networks (generator and discriminator) is also important. {\cite{grigoryev2022when}} also illustrated that generator initialization ($\mathcal{P}_{\text{init}}$) is responsible for target data modes coverage, which plays an important role in training stability. Meanwhile, discriminator initialization ($\mathcal{H}$) is responsible for an initial gradient field, which provides accurate gradient for DE-GANs training. A line of studies~\cite{yang2021one,ojha2021few,hou2022exploiting,wang2018transferring,noguchi2019image} investigated different methods to transfer GANs to novel limited datasets and report significant advantages compared to training from scratch. We divide them into three parts according to the pre-trained networks, \textit{i.e.} $\text{Generator}$, $\text{Discriminator}$, and $\text{Generator \& Discriminator}$.
\subsubsection{Generator}
Pre-training on generator is critical for DE-GANs, which provides a great initial distribution $\mathcal{P}_{\text{init}}$. {\cite{grigoryev2022when}} argued that some image-diversity priors are introduced by the pre-trained generator, and \cite{ojha2021few} believed that the generator pre-trained on large dataset can introduce latent prior to mitigate imbalance between latent and data spaces. Both of them attempted to explain the reason why pre-trained generators are effective. However, A direct fine-tuning without any regularization on the parameters often results in overfitting, since the number of parameters is significantly larger than the number of target samples. Therefore, many studies introduced different restrictions in the fine-tuning process. For instance, \cite{noguchi2019image} transferred the pre-trained network to DE-GANs by introducing scale and shift parameters to each hidden activation of the generator and updating only these parameters in the fine-tuning process. \cite{ham2020unbalanced} only pre-trained the generator of GANs using the decoder of Variational Autoencoder (VAE). The proposed unbalanced GANs outperform ordinary GANs in terms of stabilized learning, faster convergence, fewer mode collapses, and better image quality. \cite{li2020few} demonstrated that different weights of generator should have different freedom to change according to importance. Therefore, it adopted the Fisher information $F(\cdot)$ as an important measurement to penalize the updating of parameters:
\begin{equation}
	\sum_{i} F_{i}\left(\theta_{i}-\theta_{S, i}\right)^{2},\quad \text{where}\quad F=\mathbb{E}\left[-\frac{\partial^{2}}{\partial \theta_{S}^{2}} \mathcal{L}\left(X \mid \theta_{S}\right)\right],
\end{equation}
$\theta_{S}$ represents the parameters learned from the source domain, and $i$ is the index of each parameter of the model. \cite{ojha2021few} demonstrated that relative distances of latent space in DE-GANs are depraved, which is the key point of mode collapse. It hypothesized that enforcing preservation of relative pairwise distances, before and after fine-tuning, would help prevent collapse. Accordingly, they proposed a cross-domain distance consistency term for training generator in DE-GANs:
\begin{equation}
	\begin{aligned}
		\mathcal{L}_{\mathrm{dist}}\left(G_{s \rightarrow t}, G_{s}\right)=\mathbb{E}_{ z_{i} \sim p_{z}(z)} \sum_{i=i} D_{K L}\left(y_{i}^{s \rightarrow t, l} \| y_{i}^{s, l}\right),\\
		y_{i}^{s, l} =\operatorname{Softmax}\left(\left\{\operatorname{sim}\left(G_{s}^{l}\left(z_{i}\right), G_{s}^{l}\left(z_{j}\right)\right)\right\}_{\forall i \neq j}\right),\\
		y_{i}^{s \rightarrow t, l} =\operatorname{Softmax}\left(\left\{\operatorname{sim}\left(G_{s \rightarrow t}^{l}\left(z_{i}\right), G_{s \rightarrow t}^{l}\left(z_{j}\right)\right)\right\}_{\forall i \neq j}\right),
	\end{aligned}
\end{equation}
where $G_{s}$ and $G_{s \rightarrow t}$ denote pre-trained and fine-tuned generators, respectively.
\subsubsection{Discriminator}  
Discriminator initialization mitigates discriminator overfitting effectively, which provides a responsible gradient field for generator's training. Nevertheless, the naive adaptation of discriminator does not lead to promising results~\cite{sauer2021projected} since strong pre-trained features enable the discriminator to dominate the two-player game, resulting in vanishing gradients for the generator. Thus, \cite{sauer2021projected,kumari2021ensembling} adopted an image classification models trained on ImageNet as feature projection to rebuild the discriminator. They identified two key components for exploiting the full potential of pre-trained models in DE-GANs training: (i) feature pyramids to enable multi-scale feedback with multiple discriminators, and (ii) random projections to better utilize deeper layers of the pre-trained network. The experiments consistently approve the advancement of the projected GANs in terms of generative quality and convergence speed. However, this ImageNet-classification based pre-training may result in the failure \cite{kynkaanniemi2022role} of the FID metric since it is also calculated with the ImageNet pre-trained classifier. This seems to be the case, as we observed the generation results in the \cite{sauer2021projected,kumari2021ensembling}: although they have a better FID metric, the visual quality has decreased instead. Furthermore, kynkaanniemi \textit{et al.} \cite{kynkaanniemi2022role} show that FID can be significantly reduced without actually improving the quality of results, which demonstrates that a part of the observed FID improvement turns out not to be real in ImageNet pre-training discriminator. 

\subsubsection{$\text{Generator \& Discriminator}$} Pre-training of generator and discriminator affect both $\mathcal{P}_{\text{init}}$ and $\mathcal{H}$. \cite{wang2018transferring} was the first work to adopt the pre-training method to image generation. It pre-trained both the generator and discriminator on large datasets. The results show that using knowledge from pre-trained networks can shorten the convergence time and can significantly improve the quality of the generated images, especially in DE-GANs. However, ~\cite{mo2020freeze,zhao2020leveraging,robb2020few,mangla2020data,yang2021one} also illustrated that naive adaptation of generator and discriminator in DE-GANs does not lead to convincing results. These methods proposed many technologies to restrict fine-tuning processing. For instance, \cite{wang2020minegan} proposed a two-stage method. The first stage steers the latent space of the fixed generator $G$ to suitable areas for the target limited data distribution. The second stage updates the weights of the generator and discriminator via fine-tuning. According to these two stages, it effectively transfers knowledge to domains with few target images, outperforming the previous method~\cite{wang2018transferring}. \cite{mo2020freeze} introduced a simple solution: freezing the lower layers of the discriminator and fine-tuning the other layers.  \cite{zhao2020leveraging} revealed that low-level filters of both pre-trained generator and discriminator can be transferred to facilitate generation in a perceptually distinct target domain with limited training data. Furthermore, they also proposed adaptive filter modulation to adapt the transferred filters to the target domain. \cite{robb2020few} only adapted the singular values of the pre-trained weights while freezing the corresponding singular vectors. Different from the above studies that adopt the fine-tuning strategy on original networks, \cite{yang2021one} employed two lightweight modules before pre-trained generator and after pre-trained discriminator yet freeze original parameters of GANs. Concretely, an attribute adaptor is introduced into generator, through which it can retain enough prior about the balance between the latent and sampled spaces. An attribute classifier is also introduced into the well-trained discriminator backbone to avoid overfitting of the discriminator. The proposed method brings appealing results under various settings. \cite{hou2022exploiting} treated the diversity of generator and fidelity of discriminator in the source model as a kind of knowledge and proposed to improve the generation results via exploring knowledge distillation. The source model is regarded as the teacher model and the target model as the student is learned from the guidance of the teacher model. 

\begin{table*}
	\centering
\caption{FID (lower is better) metric for 10-shot generation on FFHQ-babies, FFHQ-sunglasses, and face sketches datasets (All of the results are from the reference papers).
\label{Tab:few-shot} }
	\begin{tabular}{ccccc}
		\toprule
		10-shot generation&Methods&Babies&Sunglasses&Sketches\\
		\midrule
		\multirowcell{9}{Knowledge Sharing}&
		TGAN ~\cite{wang2018transferring} & 104.79& 55.61& 53.41\\
		&TGAN+ADA~\cite{karras2020training}&102.58& 53.64 &66.99\\
		&BSA ~\cite{noguchi2019image}&140.34 &76.12 &69.32\\
		&FreezeD~\cite{mo2020freeze} & 110.92 &51.29 &46.54\\
		&MineGAN~\cite{wang2020minegan} & 98.23& 68.91 &64.34\\
		&EWC~\cite{li2020few} & 87.41& 59.73 &71.25\\
		&CDC~\cite{ojha2021few} & 74.39 &42.13& 45.67\\
		&KD-GAN~\cite{hou2022exploiting}&68.67& 34.61&35.87\\
		&GenDA~\cite{yang2021one}& $\textbf{47.05}$ & $\textbf{22.62}$ & $\textbf{31.97}$ \\
		\hline
		\multirow{2}{*}{Model Architecture}&Stylegan2~\cite{karras2020analyzing}&184.77&-&94.16\\
		&FastGAN~\cite{liu2020towards}&-&-&76.28\\
		\hline
		\multirowcell{4}{Data Augmentation\\  Regularization}&Stylegan2+DiffAug~\cite{zhao2020differentiable}&-&-&43.15\\
		&Stylegan2+ADA~\cite{karras2020training}&-&-&62.82\\
		&Stylegan2+MDL~\cite{kong2021smoothing}&93.15&-&40.02\\
		&Stylegan2+DiffAug+MDL&-&-&35.59\\
		\bottomrule
	\end{tabular}
\end{table*}

\begin{table*}[t!]
	\caption{Comparison with previous works over 100-shot and AFHQ: Training with 100 (Obama, Grumpy Cat, and Panda), 160 (AFHQ Cat), and 389 (AFHQ Dog) samples. FID (lower is better) is reported as the evaluation metric. All methods adopt the StyleGAN2 architecture except for FastGAN+DA.
		\label{Tab:simple-few-shot}}
	\small
	\centering
	\begin{tabular}{cc|ccc|cc}
		\hline\multirow{2}{*}{Few-shot generation}& 	\multirow{2}{*}{Method} & \multicolumn{3}{|c|}{ 100-shot } & \multicolumn{2}{c}{ AnimalFace } \\
		&& Obama & Grumpy Cat & Panda & Cat & Dog \\
		\hline 
		\multirowcell{4}{Knowledge Sharing}&
		Scale/shift \cite{noguchi2019image}  & 50.72 & 34.20 & 21.38 & 54.83 & 83.04 \\
		&MineGAN \cite{wang2020minegan}  & 50.63 & 34.54 & 14.84 & 54.45 & 93.03 \\
		&TGAN \cite{wang2018transferring}  & 48.73 & 34.06 & 23.20 & 52.61 & 82.38 \\
		&FreezeD \cite{mo2020freeze}  & 41.87 & 31.22 & 17.95 & 47.70 & 70.46 \\
		\hline
		{Model Architecture}
		& StyleGAN2 \cite{karras2020analyzing}  & 80.20 & 48.90 & 34.27 & 71.71 & 130.19 \\
		\hline
		\multirowcell{7}{Data Augmentation\\  Regularization}
		&TGAN + DiffAug \cite{zhao2020differentiable}  & 39.85 & 29.77 & 17.12 & 49.10 & 65.57 \\
		&StyleGAN2+DiffAug \cite{zhao2020differentiable}  & 46.87 & 27.08 & 12.06 & 42.44 & 58.85 \\
		&FastGAN+DiffAug \cite{liu2020towards}&41.05&26.65&10.03&35.11&50.66\\
		&StyleGAN2+ADA \cite{karras2020training}  & 45.69 & 26.62 & 12.90 & 40.77 & 56.83 \\
		&StyleGAN2+MDL \cite{kong2021smoothing} & 45.37 & 26.52 & -& - & - \\
		&StyleGAN2+LeCam-GAN \cite{tseng2021regularizing}  & 33.16 & 24.93 & 10.16 & 34.18 & 54.88 \\
		&StyleGAN2+GenCo \cite{cui2021genco}& 32.21&\textbf{17.79}&9.49& 30.89& 49.63 \\
		\hline
	\end{tabular}
\end{table*}
\section{Tasks, Datasets, Technologies, and Evaluation}
According to the scale of $\mathcal{I}_\text{train}$, we divide data-efficient generation into two tasks: few-shot generation and limited-data generation. Although both of them have similar settings and motivations, the employed techniques and datasets are slightly different. In this section, we conclude the characteristics of few-shot generation and limited-data generation with respect to datasets, technologies, and evaluation. 
\subsection{Few-shot Generation}
Few-shot generation is a specific data-efficient generation with very few data, for instance, 10-shot and 100-shot generation. Many datasets are adopted to evaluate the performance of few-shot generation, the most commonly used of which are FFHQ-babies, FFHQ-sunglasses, and face sketches~\cite{ojha2021few}. These datasets roughly contain 2500, 2700, and 300 images with $256 \times 256$ resolution, respectively. For 10-shot generation, 10 images from each dataset are sampled to train the model and the entire dataset is applied to measure the performance, respectively. Furthermore, some other datasets have also been used to evaluate the performance of sample few-shot generation, such as Obama, Grumpy Cat, Panda, AnimalFace Cat, and AnimalFace Dog. These datasets contain 100, 100, 100, 160, and 389 images with $256 \times 256$, respectively, and the datasets can be found at the \href{https://drive.google.com/file/d/1aAJCZbXNHyraJ6Mi13dSbe7pTyfPXha0/view}{link}.  For simple few-shot generation, all images in the dataset are used for training and measuring. In this setting, only the $\mathcal{E}_{\text {emp}}(\mathcal{H},\mathcal{P}_{\text{init}},\mathcal{I}_{\text{train}})$ needs to be minimized, which only evaluates the stability of DE-GANs and does not assess the diversity and generalization of the generated samples. Indeed, current state-of-the-art methods in simple few-shot generation tasks tend to remember data samples, which dramatically limits the scope of DE-GANs. We expect to generate more information beyond the training sample rather than simply remembering the current sample.

Few-shot generation is a hard task due to the employed extremely small amount of data. There are two characteristics in the few-shot generation: (i) Most studies improve the performance through introducing knowledge from large-scale datasets. In general, freezing more pre-trained layers of networks leads to better performance. For instance, GenDA \cite{yang2021one} imported two lightweight modules before pre-trained generator and after pre-trained discriminator yet freeze original parameters of GANs, which has the state-of-the-art performance on the 10-shot generation of various datasets (Table \ref{Tab:few-shot}). (ii) Compared to limited-data generation, few-shot generation focuses more on the quality of the generated images. In general, some technologies of stricter limits on diversity are adopted in the few-shot generation. For instance, \cite{yang2021one} proposed a latent diversity constraint strategy during training, which truncates the latent distribution with a strength factor $\beta$ as: $\mathbf{z}^{\prime}=A(\beta \mathbf{z}+(1-\beta) \overline{\mathbf{z}}),$
where $\overline{\mathbf{z}}$ indicates the mean code. ~\cite{kong2021smoothing} smoothed the latent space of GANs with mixup-based distance learning, which improves the smoothness and interpolation, but also reduces diversity of the latent space. Table~\ref{Tab:few-shot} illustrates the FID metric for 10-shot generation on popular datasets. Furthermore, Table \ref{Tab:simple-few-shot} illustrates the FID \cite{heusel2017gans} metric of various methods for simple few-shot generation on popular datasets. Among them, StyleGAN2 with GenCo has achieved the best performance.

\begin{table*}
	\centering
	\caption{FID (lower is better) metric for limited-data generation on FFHQ-5K, FFHQ-1K, and CIFAR-10 ($10\%$) datasets (All of the results are from the reference papers).
		\label{Tab:partial-data}}
	\begin{tabular}{cccc}
		\toprule
		Limited-data generation&FFHQ-5K(256 $\times$ 256)&FFHQ-1K&CIFAR-10 ($10\%$)\\
		\midrule
		Stylegan2~\cite{karras2020analyzing}&49.68&100.16&36.02\\
		Stylegan2+GenCo~\cite{cui2021genco}&27.96&65.31& 27.16\\
		Stylegan2+ADA~\cite{karras2020training}&10.96&21.29&6.72\\
		Stylegan2+ADA+$R_{LC}$~\cite{tseng2021regularizing}&-&21.7&6.56\\
		Stylegan2+ADA+Insgen~\cite{yang2021data}&-&19.58&-\\
		Stylegan2+ADA+APA~\cite{jiang2021deceive}&$\textbf{8.38}$&$\textbf{18.89}$&-\\
		\bottomrule
	\end{tabular}
	
\end{table*}
\subsection{Limited-data Generation}
Compared to few-shot generation, limited-data generation employs more data, such as 1K or 5K. The most commonly used datasets are FFHQ and CIFAR-10. FFHQ contains 70K high-resolution images of human faces. For the experiments of
limited data generation, the images are resized to $256 \times 256$. Furthermore, a subset of dataset by randomly sampling with 1K (FFHQ-1K) and 5K (FFHQ-5K) images is collected to train. Regardless of the number of the training data, the FID \cite{heusel2017gans} metric is calculated between 50K fake images and all 70K real images, evaluating the performance in terms of both reality and diversity. Similar, CIFAR-10 contains 50K images with $32 \times 32$. For the experiments of
limited data generation, a subset of CIFAR-10 by randomly sampling with $10\%$ images is collected to train. FID is calculated between 50K fake
images and all 50K real images. Usually, non-pre-training methods, such as data augmentation and regularization, are sufficient for stability training in the limited-data generation. Table \ref{Tab:partial-data} illustrates FID metric for some popular and advanced methods on various datasets. From the Table \ref{Tab:partial-data}, ADA~\cite{karras2020training} is compatible with all other methods and APA~\cite{jiang2021deceive} building on ADA acquires state-of-the-art performance. 


\section{Challenges and Opportunities}

The current technologies in DE-GANs have achieved stable training and good performance. Here, we attempt to highlight the existing problems and the open issues that would inspire future research. 

\textit{New Metrics:} FID between generated and entire distributions is invariably used in the metrics of DE-GANs. However, ImageNet-classification based pre-training~\cite{sauer2021projected,kumari2021ensembling} may lead to the leakage of FID to training and mismatching between FID and image reality. Furthermore, by choosing a subset from the slightly larger set of generated images, FID can be significantly reduced without improving the quality of results \cite{kynkaanniemi2022role}. Both of them suggest caution against over-interpreting FID improvements on reality. In addition to reality, smoothness and interpolation of latent space and image diversity are critical for DE-GANs. Future research can try to design a new metric that can comprehensively consider the image quality, image diversity, latent space smoothness, and latent space interpolation.

\textit{Data Selection:} 
The current data selection strategy includes random sampling and high-density sampling. Random sampling does not take into account the adaptability of GANs to the data density, and high-density sampling decreases the difficulty of the task and the diversity of data. Future research can try to propose a data selection strategy that balances the difficulty of the task and the diversity of data.

\textit{Universal Technology Between Few-shot and Limited-data Generation:} 
With the exception of data augmentation, other technologies are not universal and cannot be applied simultaneously to both few-shot and limited-data generation. The key point is that smaller amount of data leads to tighter optimization restrictions, which makes the methods of few-shot generation have a deficient ability to fit the distribution in the limited-data generation task. In contrast, methods of limited-data generation do not have the stability of training in the few-shot generation task. Some universal or adaptive methods can try to be proposed In the future.

\textit{Data Biased Generation:} An important but unexplored task, data-biased generation, should be researched in the future. Data biased generation has been defined as: we expect to train DE-GANs with the biased training dataset $\mathcal{I_{\text{train}}}$ to fit the unbiased expected distribution $\mathcal{P}_\text{exp}(\mathcal{I})$. How to debias the generation with limited data will be a challenging task.

\section{Conclusion}
Data-Efficient GANs have received attention due to the burden of collecting large-scale high-resolution data in industrial applications. Recently, many techniques involving DE-GANs training have been reported. However, a comprehensive and systematic survey DE-GANs remains vacant. In this paper, we provide a formal description of data-efficient GANs and divide them into two tasks based on the size of the training data. We then point out that the degradation of DE-GANs comes from the gap between two data distributions and the difficulties of GANs training due to few samples. Understanding the degradation of DE-GANs helps categorize different works into Data Selection, GANs Optimization, and Knowledge Sharing. In addition, we have summarized the performance of some popular methods and made recommendations for the selection of datasets based on two tasks. Finally, we point out the issues worthy of attention in the future of DE-GANs.

\small
\bibliographystyle{named}
\bibliography{ijcai22}

\end{document}